\documentclass[sigconf,
                nonacm,
        ]{acmart}
\usepackage{amsmath}
\usepackage{graphicx}
\usepackage{amsfonts}
\usepackage{amsthm}
\usepackage{booktabs}
\usepackage[ruled,vlined,linesnumbered]{algorithm2e}
\usepackage{dsfont}
\usepackage{pifont}
\newcommand{\cmark}{\ding{51}}%
\newcommand{\xmark}{\ding{55}}%
\AtBeginDocument{%
  }

\setcopyright{acmlicensed}
\copyrightyear{2024}
\acmYear{2024}
\acmDOI{XXXXXXX.XXXXXXX}

\acmConference[ICAIF '24]{International Conference on AI in Finance}{November 14--16,
  2024}{Brooklyn, NY}
\acmISBN{978-1-4503-XXXX-X/18/06}

\begin{document}

%%
%% The "title" command has an optional parameter,
%% allowing the author to define a "short title" to be used in page headers.

% \title[Contrastive Learning of Asset Embeddings]{Contrastive Representation Learning from Time Series for Financial Asset Embeddings}
\title[Contrastive Learning of Asset Embeddings]{Contrastive Learning of Asset Embeddings\\from Financial Time Series}

%%
%% The "author" command and its associated commands are used to define
%% the authors and their affiliations.
%% Of note is the shared affiliation of the first two authors, and the
%% "authornote" and "authornotemark" commands
%% used to denote shared contribution to the research.
\author{Rian Dolphin}
\orcid{0000-0002-5607-9948}
\affiliation{%
  \institution{School of Computer Science, University College Dublin}
  \city{Dublin}
  \country{Ireland}
}
\email{rian.dolphin@ucdconnect.ie}
% \email{dolphrian@gmail.com}

\author{Barry Smyth}
\orcid{0000-0003-0962-3362}
\affiliation{%
  \institution{School of Computer Science, University College Dublin}
  \city{Dublin}
  \country{Ireland}
}
\email{barry.smyth@ucd.ie}

\author{Ruihai Dong}
\orcid{0000-0002-2509-1370}
\affiliation{%
  \institution{School of Computer Science, University College Dublin}
  \city{Dublin}
  \country{Ireland}
}
\email{ruihai.dong@ucd.ie}

%%
%% By default, the full list of authors will be used in the page
%% headers. Often, this list is too long, and will overlap
%% other information printed in the page headers. This command allows
%% the author to define a more concise list
%% of authors' names for this purpose.
\renewcommand{\shortauthors}{Dolphin et al.}

%%
%% The abstract is a short summary of the work to be presented in the
%% article.
\begin{abstract}
Representation learning has emerged as a powerful paradigm for extracting valuable latent features from complex, high-dimensional data. In financial domains, learning informative representations for assets can be used for tasks like sector classification, and risk management. However, the complex and stochastic nature of financial markets poses unique challenges. We propose a novel contrastive learning framework to generate asset embeddings from financial time series data. Our approach leverages the similarity of asset returns over many subwindows to generate informative positive and negative samples, using a statistical sampling strategy based on hypothesis testing to address the noisy nature of financial data. We explore various contrastive loss functions that capture the relationships between assets in different ways to learn a discriminative representation space. Experiments on real-world datasets demonstrate the effectiveness of the learned asset embeddings on benchmark industry classification and portfolio optimization tasks. In each case our novel approaches significantly outperform existing baselines highlighting the potential for contrastive learning to capture meaningful and actionable relationships in financial data.

\end{abstract}

%%
%% The code below is generated by the tool at http://dl.acm.org/ccs.cfm.
%% Please copy and paste the code instead of the example below.
%%


%%
%% Keywords. The author(s) should pick words that accurately describe
%% the work being presented. Separate the keywords with commas.
%% A "teaser" image appears between the author and affiliation
%% information and the body of the document, and typically spans the
%% page.
% \begin{teaserfigure}
%   \includegraphics[width=\textwidth]{sampleteaser}
%   \caption{Seattle Mariners at Spring Training, 2010.}
%   \Description{Enjoying the baseball game from the third-base
%   seats. Ichiro Suzuki preparing to bat.}
%   \label{fig:teaser}
% \end{teaserfigure}

% \received{20 February 2007}
% \received[revised]{12 March 2009}
% \received[accepted]{5 June 2009}

%%
%% This command processes the author and affiliation and title
%% information and builds the first part of the formatted document.
\maketitle

\section{Introduction}
Representation learning has emerged as a powerful paradigm for extracting valuable latent features from complex and high-dimensional data across various domains, such as computer vision~\cite{chen2020simple}, natural language processing~\cite{mikolov2013efficient}, and graph mining~\cite{kipf2016semi}. In the field of finance, learning informative representations of financial assets is of use for a wide range of applications, including portfolio optimization, risk management, and sector classification. However, the complex and stochastic nature of financial markets poses unique challenges for representation learning techniques, as the underlying relationships between assets are often non-linear, time-varying, and influenced by a multitude of factors~\cite{cont2001empirical}.

Traditional approaches to modeling financial assets rely on handcrafted features or statistical measures, such as historical returns, volatility, and correlation~\cite{markowitz1952portfolio}. While these methods have been widely used in practice, they often struggle to capture the intricate dependencies and dynamics present in market data~\cite{cont2001empirical}. Moreover, the increasing availability of high-frequency trading data and the growing complexity of financial instruments call for more sophisticated and data-driven approaches to representation learning in finance~\cite{sezer2020financial}.

Recent advancements in deep learning and self-supervised learning have shown promising results in learning meaningful representations from raw data in various domains~\cite{chen2020simple,devlin2018bert}. In particular, contrastive learning has emerged as a powerful framework for learning representations by maximizing the similarity between positive pairs of samples while minimizing the similarity between negative pairs~\cite{oord2018representation}. Contrastive learning has been successfully applied to learn representations from images~\cite{chen2020simple}, text~\cite{logeswaran2018efficient}, and graphs~\cite{you2020graph}, demonstrating its effectiveness in capturing the underlying structure and relationships in complex data.

In this paper, we propose a novel contrastive learning framework for learning asset embeddings from financial time series data. Our approach aims to capture the complex relationships and similarities between assets by leveraging their returns similarity over rolling subwindows through time. We introduce a statistical sampling strategy based on a hypothesis test of proportions to generate informative positive and negative samples for the contrastive learning process. This sampling strategy addresses the challenges associated with the noisy and stochastic nature of financial data, enabling the learning of more robust and meaningful asset embeddings.

We explore various contrastive loss functions that weight the relationships between the anchor asset and the positive and negative samples in different ways. These loss functions are designed to pull the embeddings of similar assets closer together while pushing the embeddings of dissimilar assets apart, thereby learning a discriminative and informative representation space. We evaluate the effectiveness of our learned asset embeddings on two downstream financial tasks: industry sector classification and a na\"ive risk hedging experiment. Our experiments on real-world financial datasets demonstrate that our approach achieves state-of-the-art performance in these tasks, outperforming traditional methods and showcasing the practical value of our contrastive learning framework.

The main contributions of this paper can be summarized as follows:

\begin{itemize}
    \item We propose a novel contrastive learning framework for learning asset embeddings from financial time series data, leveraging their returns similarity over rolling subwindows.
    \item We introduce a statistical sampling strategy based on a hypothesis test of proportions to generate informative positive and negative samples for the contrastive learning process, addressing the challenges associated with the noisy and stochastic nature of financial data.
    \item We explore various contrastive loss functions that capture the relationships between the anchor asset and the positive and negative samples in different ways, learning a discriminative and informative representation space.
    \item We conduct rigorous experiments on real-world financial datasets and demonstrate the effectiveness of our learned asset embeddings in downstream tasks such as industry sector classification and risk hedging, outperforming traditional methods and showcasing the practical value of our approach.
\end{itemize}

The remainder of this paper is organized as follows. Section \ref{sec:related_work} provides an overview of related work on representation learning in finance and contrastive learning. Section \ref{sec:methodology} describes our proposed contrastive learning framework, including the statistical sampling strategy and the explored contrastive loss functions. Section \ref{sec:experiments} presents the experimental setup and results on industry sector classification and risk hedging tasks. Finally, Section \ref{sec:conclusion} concludes the paper and discusses future research directions.

\section{Related Work}\label{sec:related_work}

Representation learning has emerged as a powerful paradigm for extracting meaningful features from complex and high-dimensional data across various domains, such as computer vision~\cite{chen2020simple}, natural language processing~\cite{mikolov2013efficient}, and graph mining~\cite{kipf2016semi}. In the field of finance, learning informative representations of financial assets is of paramount importance for a wide range of applications, including portfolio optimization, risk management, and sector classification~\cite{dolphin2022stock}.

Traditional approaches to representation learning in finance have relied on extracting handcrafted features from financial time series data, such as statistical measures and technical indicators~\cite{tsay2005analysis,murphy1999technical}. Econometric models, like the Capital Asset Pricing Model (CAPM)~\cite{sharpe1964capital} and the Arbitrage Pricing Theory~\cite{ross1976arbitrage}, have also been used to capture the relationship between asset returns and various risk factors. Dimensionality reduction techniques, such as factor analysis, have also been applied to financial time series data to extract low-dimensional representations~\cite{tsay2005analysis}.

In recent years, deep learning techniques have gained significant attention in the financial domain. Recurrent Neural Networks (RNNs), particularly Long Short-Term Memory (LSTM) networks~\cite{hochreiter1997long}, have been widely employed to model the temporal dependencies and capture the long-term patterns in financial time series data~\cite{bao2017deep}. Furthermore, complex deep learning techniques that have become popular in NLP and computer vision, like transformers and diffusion models, have also been applied to financial market data~\cite{voigt2024assessment, daiya2024diffstock}.

Self-supervised learning techniques have also emerged as a promising approach for learning meaningful representations from financial time series data. These methods aim to learn representations by solving pretext tasks that do not require explicit labels. For example, \citet{sarmah2022learning} proposed a framework for learning asset embeddings by applying node2vec~\cite{grover2016node2vec} to a graph derived from returns correlations, while \citet{dolphin2022machine,dolphin2023iccbr} also leverage pairwise similarity between returns time series to learn asset embeddings. Beyond time series, \citet{gabaix2023asset} and \citet{satone2021fund2vec} introduced approaches for learning representations from holdings data, for assets and funds respectively. Furthermore, \citet{ito2020embedding} leverage pre-trained language models to generate representations from public textual filings like 10-K reports.

Contrastive learning, a popular approach within self-supervised learning, has shown remarkable success in learning meaningful representations from complex data in various domains. In computer vision, techniques like SimCLR~\cite{chen2020simple} have achieved state-of-the-art performance on image classification tasks by learning representations that maximize the similarity between positive pairs of images while minimizing the similarity between negative pairs. In NLP, contrastive learning has been applied to learn word embeddings~\cite{mikolov2013efficient} and sentence representations~\cite{logeswaran2018efficient}.

Despite the success of contrastive learning in other domains, its application to financial time series data remains relatively unexplored. The complex and stochastic nature of financial markets poses unique challenges for representation learning techniques, as the underlying relationships between assets are often non-linear, time-varying, and influenced by a multitude of factors~\cite{cont2001empirical}. 

In this paper, we propose a novel contrastive learning framework for learning asset embeddings from financial time series data. Our approach aims to capture the complex relationships and similarities between assets by leveraging the co-occurrence of assets in their returns similarity. We introduce a statistical sampling strategy based on a hypothesis test of proportions to generate informative positive and negative samples for the contrastive learning process, addressing the challenges associated with the noisy and stochastic nature of financial data. We explore various contrastive loss functions that capture the relationships between the anchor asset and the positive and negative samples in different ways, learning a discriminative and informative representation space.

Our work contributes to the growing body of research on representation learning in finance and advances the application of contrastive learning techniques to financial time series data. By learning meaningful and informative asset embeddings, our approach has the potential to improve the performance of downstream financial tasks and provide new insights into the complex dynamics of financial markets.

\section{Methodology}\label{sec:methodology}

In this section, we present our proposed contrastive learning framework for learning asset embeddings from financial time series data. We first introduce the problem setting and notation, followed by a description of our statistical sampling strategy for generating positive and negative samples. We then present the various contrastive loss functions explored in our framework and discuss the training and optimization procedure.

\subsection{Problem Setting and Notation}
Let $\mathcal{A} = \{a_1, a_2, \dots, a_N\}$ be a set of $N$ financial assets, where each asset $a_i$ is associated with a time series of daily returns $\mathbf{r}_{a_i} = \{r_1^{a_i}, r_2^{a_i}, \dots, r_T^{a_i}\}$ over the same period of $T$ time steps. The return $r_t^{a_i}$ of asset $a_i$ at time $t$ is calculated as:

\begin{equation}
    r_t^{a_i} = \frac{p_t^{a_i} - p_{t-1}^{a_i}}{p_{t-1}^{a_i}},
\end{equation}
where $p_t^{a_i}$ denotes the price of asset $a_i$ at time $t$.

Our goal is to learn an embedding function $f: \mathcal{A} \rightarrow \mathbb{R}^d$ that maps each asset $a_i$ to a $d$-dimensional embedding vector $\mathbf{e}_i \in \mathbb{R}^d$, such that the embeddings capture the underlying relationships and similarities between the assets. We denote the embedding matrix as $\mathbf{E} \in \mathbb{R}^{N \times d}$, where each row corresponds to the embedding vector of an asset.

\subsection{Generating Positive and Negative Samples}
A key component of our contrastive learning framework is the generation of positive and negative sampling distributions for each asset. We propose a statistical sampling strategy based on the co-occurrence of assets in their returns similarity over a sliding window.

Given an asset $a_i$, we consider a sliding window of length $w$ and stride $s$ over its returns time series. For each window, we compute the pairwise similarities between the returns subsequence of $a_i$ and the returns subsequences of all other assets within the same window. We use a similarity function $sim(\cdot,\cdot)$ to measure the similarity between two returns subsequences. The choice of the similarity function can be adapted to the specific characteristics of the financial time series data, such as Pearson correlation, dynamic time warping~\cite{berndt1994using}, or domain-specific measures~\cite{dolphin2021measuring}.

\begin{figure}
    \centering
    \includegraphics[width=0.47\textwidth]{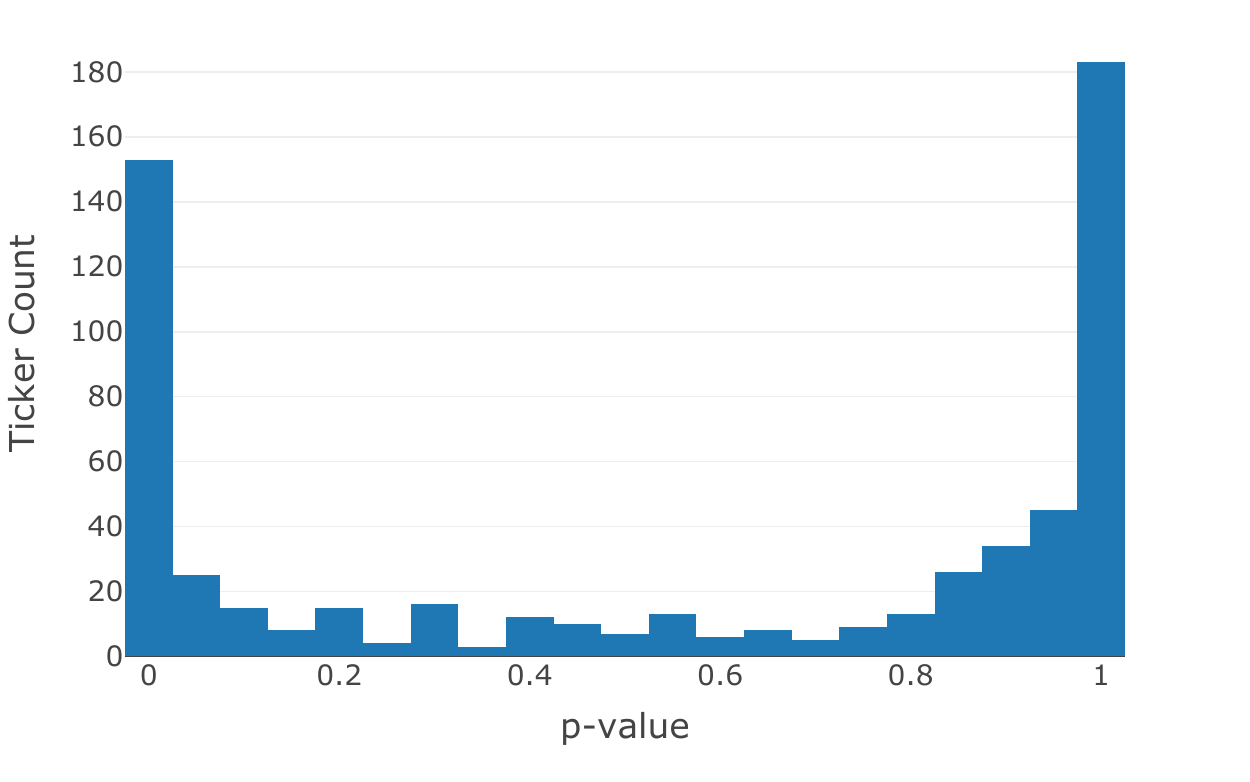}
    \caption{Observed distribution of $p$-values for proportion test.}
    \label{fig:contrastive:p_value_distribution}
\end{figure}

\subsubsection{Co-occurrence Matrix}
We then construct a co-occurrence count matrix $\mathbf{C} \in \mathbb{N}^{N \times N}$, where each entry $\mathbf{C}_{i,j}$ represents the number of sliding windows in which asset $a_j$ appears among the top-$k$ most similar assets to $a_i$ across all sliding windows. Formally,
\begin{equation}
\mathbf{C}_{i,j} = \sum_{t=1}^{\left\lfloor\frac{T-w}{s}\right\rfloor + 1} \mathds{1}\left[j \in \mathrm{topk}\left(\mathbf{s}_{a_i,t}, k\right)\right],
\end{equation}
where
\begin{itemize}
    \item $\mathbf{s}_{a_i,t}$ is the vector of similarities between asset $a_i$ to all other assets, within the sliding window of length $w$ starting at time $t_s=(t-1) \cdot s + 1$, and is given by 
    \begin{multline}
        \hspace{1cm}\mathbf{s}_{a_i,t} = \Big[sim\left(\mathbf{r}_{a_i}[t_s:t_s+w], \mathbf{r}_{a_j}[t_s:t_s+w]\right) \\
        \bigm| j \in \{1, \dots, N\}, j \neq i \Big]
    \end{multline}
    The introduction of $t_s$ for indexing is to ensure that the sliding window moves by the specified stride $s$ at each step.

    \item $\mathrm{topk}(\mathbf{x}, k)$ returns the indices of the top-$k$ largest elements in a vector $\mathbf{x}$.
    \item $\mathds{1}[\cdot]$ is the indicator function.
    \item $\mathbf{r}_{a_i}[q:q+r]$ is the subsequence taken from $\mathbf{r}_{a_i}$ of length $r$ starting at index $q$
\end{itemize}

In this way, $\mathbf{C}_{i,j}$ denotes the number of sub windows where asset $a_j$ was among the top-$k$ most similar assets to $a_i$ over all sliding windows considered.

\subsubsection{Sampling Strategy}
To generate positive and negative samples for an asset $a_i$, we propose a hypothesis testing approach based on the co-occurrence matrix $\mathbf{C}$. We consider the null hypothesis $H_0: p_{i,j} \leq p_0$, where $p_{i,j}$ is the probability of asset $a_j$ co-occurring with asset $a_i$, and $p_0$ is the expected co-occurrence probability under the assumption of equal likelihood, i.e., $p_0 = \frac{1}{N}$.

We compute the p-value for each pair of assets $(a_i, a_j)$ using the test statistic:

\begin{equation}
    z_{i,j} = \frac{\hat{p}_{i,j} - p_0}{\sqrt{\frac{p_0(1-p_0)}{n_i}}},
\end{equation}
where $\hat{p}_{i,j} = \frac{\mathbf{C}_{i,j}}{n_i}$ is the empirical co-occurrence probability, and $n_i = \sum_{j=1}^N \mathbf{C}_{i,j}$ is the total number of co-occurrences for asset $a_i$. The p-value is then given by $\mathtt{p}(a_i, a_j) = 1-\Phi(z_{i,j})$ where $\Phi(\cdot)$ is the normal cumulative distribution function. The observed distribution of p-values is shown in Figure \ref{fig:contrastive:p_value_distribution}. We see that the tails of the distribution, indicating asset co-occurrence significantly below and above random chance, are highly populated.

To generate positive samples for asset $a_i$, we only sample assets $a_j$ that have a p-value below a threshold $\alpha_p$, indicating a significantly higher observed co-occurrence count than expected by random chance. The sampling distribution is constructed such that the selection probability is inversely proportional to p-value. Formally, the positive sampling distribution is defined as:
\begin{equation}
\mathbf{P}_{i,j} =
\begin{cases}
\frac{\displaystyle\mathbf{C}_{i,j}}{\displaystyle\sum_{\{l ~:~ \texttt{p}(a_i,a_l) < \alpha^+\}} \mathbf{C}_{i,l}} & \text{if } \texttt{p}(a_i,a_j) < \alpha^+ \\[30pt]
0 & \text{if } \texttt{p}(a_i,a_j) > \alpha^+
\end{cases}
\end{equation}
where $\mathbf{P}_{i,j}$ is the probability that $a_j$ is drawn as a positive sample for $a_i$. From this distribution we can then draw a set of positive samples for asset $a_i$, which we denote $\mathcal{P}_i$.

For negative samples, we select assets $a_j$ that have a p-value above a threshold $\alpha_n$, indicating a significantly lower co-occurrence count than expected by random chance. In a similar way to $\mathbf{P}$ we define the negative sampling distribution as:
\begin{equation}\label{eqn:contrastive:negative_sampling_formula}
\mathbf{N}_{i,j} =
\begin{cases}
\frac{\displaystyle\max_{\substack{1\leq m\leq N \\ m\neq i}}\mathbf{C}_{i,m} - \mathbf{C}_{i,j}}{\displaystyle\sum_{\{l ~:~ \texttt{p}(a_i,a_l) > \alpha^-\}} \mathbf{C}_{i,l}} & \text{if } \texttt{p}(a_i,a_j) > \alpha^- \\[30pt]
0 & \text{if } \texttt{p}(a_i,a_j) < \alpha^-
\end{cases}
\end{equation}
In this case, assets with higher p-values (lower co-occurrence) are sampled more often. $\mathbf{N}_{i,j}$ denotes the probability that $a_j$ is drawn as a positive sample for $a_i$ and the set of negative samples for asset $a_i$ is given by $\mathcal{N}_i$.

\begin{figure}
    \centering
    \includegraphics[width=0.47\textwidth]{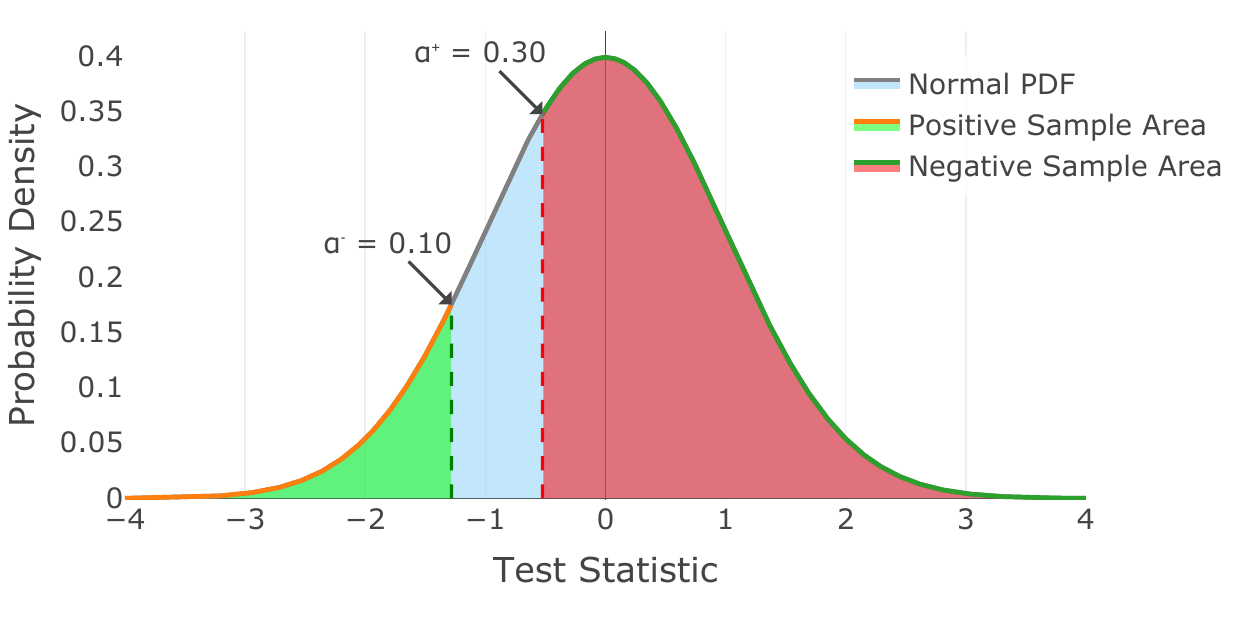}
    \caption{Example sampling regions based on the computed test statistic. The negative sampling region is much larger than the positive sampling region to promote diverse negatives and prevent degenerate solutions.}
    \label{fig:contrastive:sampling_tails}
\end{figure}

To create a diverse set of negative samples, necessary in contrastive learning to prevent a degenerate solution~\cite{robinson2020contrastive}, we set $\alpha_n$ to be more conservative than $\alpha_p$, allowing for the inclusion of moderately dissimilar assets in the negative sets. See Figure \ref{fig:contrastive:sampling_tails} for an illustration of the respective sampling regions. The red region encompassing $z$ values of $0$ indicates that assets co-occurring at the level of random chance have non-zero probability in the negative sampling distribution.

\subsection{Contrastive Loss Functions}
We explore three loss functions to learn the asset embeddings by maximizing the similarity between an anchor asset and its positive samples while minimizing the similarity between the anchor asset and its negative samples. The first does this at the level of individual positive and negative samples, while the second aggregates positive and negative sample embeddings before computing loss.

\subsubsection{Individual Sigmoid Loss}
The individual sigmoid loss considers each positive and negative sample independently and aims to maximize the similarity between the anchor asset and each positive sample while minimizing the similarity between the anchor asset and each negative sample. The loss function is defined as:

\begin{multline}
    \mathcal{L}_{\text{ind}}(a_i, \mathcal{P}_i, \mathcal{N}_i) = -\frac{1}{|\mathcal{P}_i|}\sum_{a_j \in \mathcal{P}_i} \log \sigma(\mathbf{e}_i^\top \mathbf{e}_j) \\
    - \frac{1}{|\mathcal{N}_i|}\sum_{a_j \in \mathcal{N}_i} \log(1 - \sigma(\mathbf{e}_i^\top \mathbf{e}_j)),
\end{multline}

where $\sigma(\cdot)$ is the sigmoid function, and $\mathbf{e}_i$ and $\mathbf{e}_j$ are the embedding vectors of assets $a_i$ and $a_j$, respectively.

\subsubsection{Aggregate Sigmoid Loss}
The aggregate sigmoid loss considers the average embedding of the positive samples and the average embedding of the negative samples. The loss function is defined as:

\begin{multline}
    \mathcal{L}_{\text{agg}}(a_i, \mathcal{P}_i, \mathcal{N}_i) = -\log \sigma\left(\mathbf{e}_i^\top \frac{1}{|\mathcal{P}_i|}\sum_{a_j \in \mathcal{P}_i} \mathbf{e}_j\right) \\
    - \log\left(1 - \sigma\left(\mathbf{e}_i^\top \frac{1}{|\mathcal{N}_i|}\sum_{a_j \in \mathcal{N}_i} \mathbf{e}_j\right)\right).
\end{multline}

\subsubsection{Hybrid Sigmoid-Softmax Loss}
The hybrid sigmoid-softmax loss combines the individual sigmoid loss for positive samples with a softmax loss that contrasts the similarity between the anchor asset and the average positive embedding against the similarities between the anchor asset and each individual negative sample. The loss function is defined as:

\begin{multline}
    \mathcal{L}_{\text{hybrid}}(a_i, \mathcal{P}_i, \mathcal{N}_i) = -\frac{1}{|\mathcal{P}_i|}\sum_{a_j \in \mathcal{P}_i} \log \sigma(\mathbf{e}_i^\top \mathbf{e}_j) \\
    - \log\frac{\exp\left(\mathbf{e}_i^\top \frac{1}{|\mathcal{P}_i|}\sum_{a_j \in \mathcal{P}_i} \mathbf{e}_j\right)}{\exp\left(\mathbf{e}_i^\top \frac{1}{|\mathcal{P}_i|}\sum_{a_j \in \mathcal{P}_i} \mathbf{e}_j\right) + \sum_{a_j \in \mathcal{N}_i} \exp(\mathbf{e}_i^\top \mathbf{e}_j)}.
\end{multline}

\subsection{Training and Optimization}
We train the asset embeddings using mini-batch stochastic gradient descent with the Adam optimizer~\cite{kingma2014adam}. In each training iteration, we sample a batch of anchor assets and their corresponding positive and negative samples, and compute the contrastive loss for each anchor asset using one of the loss functions described above. The gradients are then computed and used to update the embedding matrix $\mathbf{E}$.

To prevent overfitting and encourage the learned embeddings to be uniformly distributed on the unit hypersphere, we apply a custom regularization to the embedding vectors after each update step:

\begin{equation}
    \mathcal{L}_{\text{reg}} = \lambda \sum_{i=1}^{N} \left( \left\lVert \mathbf{e}_i \right\rVert_2^2 - 1 \right)^2
\end{equation}
where $\lambda=0.001$ is the regularization parameter.

We also employ a learning rate scheduler that reduces the learning rate by a factor of 0.8 when the positive sample portion of the loss plateaus for a specified number of epochs. This allows the model to fine-tune the embeddings in the later stages of training and converge to a better optimum.
The training process is summarized in Algorithm \ref{alg:training}.

\begin{algorithm}[t]
\SetAlgoLined
\KwIn{Asset returns $\{\mathbf{r}_{a_i}\}_{i=1}^N$, embedding dimension $d$, batch size $B$, number of epochs $E$, learning rate $\eta$, positive threshold $\alpha_p$, negative threshold $\alpha_n$}
\KwOut{Learned asset embeddings $\mathbf{E}$}
Initialize the embedding matrix $\mathbf{E} \in \mathbb{R}^{N \times d}$ randomly\;
Compute the co-occurrence matrix $\mathbf{C}$ based on the asset returns $\{\mathbf{r}_{a_i}\}_{i=1}^N$\;
\For{$epoch = 1, 2, \dots, E$}{
    \For{$batch = 1, 2, \dots, \lceil\frac{N}{B}\rceil$}{
        Sample a batch of anchor assets $\{a_i\}_{i=1}^B$\;
        \For{$i = 1, 2, \dots, B$}{
            Generate positive samples $\mathcal{P}_i$ based on $\mathbf{C}$ and $\alpha_p$\;
            Generate negative samples $\mathcal{N}_i$ based on $\mathbf{C}$ and $\alpha_n$\;
            Compute the contrastive loss $\mathcal{L}(a_i, \mathcal{P}_i, \mathcal{N}_i)$\;
        }
        Compute the average contrastive loss over the batch\;
        Update the embedding matrix $\mathbf{E}$ using Adam optimizer with learning rate $\eta$\;
        Apply $\ell_2$ normalization to the embedding vectors\;
    }
    Adjust the learning rate based on the validation loss\;
}
\caption{Training procedure for learning asset embeddings using contrastive learning}
\label{alg:training}
\end{algorithm}

\section{Experimental Results}\label{sec:experiments}
In this section, we present the results obtained when using our contrastive learning approach on two important tasks: industry sector classification and portfolio optimization. We compare our approach with traditional methods to demonstrate that our learned embeddings offer significant performance improvements which suggests that they are better able to capture meaningful relationships between financial assets than alternative approaches.

\subsection{Dataset and Experimental Setup}
For this evaluation we use a publicly available, real-world financial dataset consisting of daily returns for a universe of 611 U.S. stocks from 2000 to 2018\footnote{This dataset has been collected by the authors and can be found at \href{https://github.com/rian-dolphin/stock-embeddings}{https://github.com/rian-dolphin/stock-embeddings}.}. This dataset includes the industry sector labels for each stock, which are used for evaluating the sector classification task. The sector labels are based on the Global Industry Classification Standard (GICS)~\cite{GICS2020}, which categorizes companies into 11 sectors based on their principal business activities.

For the contrastive learning framework, we set the embedding dimension to $d=16$, the sliding window length to $w=22$ (corresponding to the number of trading days in a month), the stride to $s=5$ (trading days in a week), and the number of top-$k$ similar assets to $k=5$. We use the Pearson correlation coefficient as the similarity measure between returns subsequences. The positive and negative thresholds are set to $\alpha_p=0.05$ and $\alpha_n=0.3$, respectively and we train the embeddings for 30 epochs using the Adam optimizer with a learning rate of 0.001 and a batch size of 128.

\subsection{Industry Sector Classification}

\begin{table}
\centering
\begin{tabular}{p{4cm}cc}
\toprule
\textbf{TSC Baseline} & \textbf{F1} & \textbf{Accuracy} \\
\midrule
Catch22 & 0.31 & 35\% \\
Contractable BOSS & 0.37 & 39\% \\
RBOSS & 0.45 & 42\% \\
Shapelet & 0.45 & 42\% \\
Shapelet Transform & 0.40 & 46\% \\
WEASEL & 0.47 & 47\% \\
MUSE & 0.51 & 54\% \\
TS Forest Classifier & 0.53 & 55\% \\
Canonical Interval Forest & 0.52 & 56\% \\
Arsenal & 0.53 & 58\% \\
\citet{dolphin2022stock}$\textsuperscript{*}$ & 0.60 & 60\% \\
\citet{sarmah2022learning}$\textsuperscript{*}$ & 0.61 & 61\% \\
\citet{dolphin2023iccbr}$\textsuperscript{*}$ & 0.66 & 65\% \\
\midrule
\textbf{Proposed Contrastive} & & \\
\midrule
Individual Sigmoid & 0.68 & 68\% \\
Aggregate Sigmoid & 0.49 & 49\% \\
Sigmoid-Softmax & \textbf{0.69} & \textbf{69\%} \\
\bottomrule
\end{tabular}
\caption{Sector classification results for the proposed contrastive loss functions and baselines. * indicates a best-effort implementation by the authors.}
\label{tab:contrastive:sector_results}
\end{table}

To evaluate the quality of the learned asset embeddings, we first consider the task of industry sector classification. This task involves assigning companies into industry sectors (Technology, Health Care, etc.) and was chosen for two reasons. Firstly, it is frequently used as a benchmarking evaluation task~\cite{sarmah2022learning,dolphin2023iccbr}. Secondly, it is crucial to real-world financial workflows like indentifying peers and competitors within a given industry, and deciding on the constituent assets of sector-specific Exchange Traded Funds (ETFs). In fact, approximately 30\% of publications in the top-three finance journals rely on industry sector classification schemes~\cite{weiner2005impact}, further indicating their importance. 

We use the learned embeddings as input features to a support vector machine (SVM) classifier and perform 5-fold cross-validation.
Table \ref{tab:contrastive:sector_results} presents the sector classification results, in terms of the average F1-score and accuracy across the folds, for our proposed contrastive learning approach with different loss functions, along with several baseline methods. The baseline methods include traditional time series classification methods as well as state-of-the-art task specific baselines~\cite{sarmah2022learning,dolphin2022stock}.

The results show that our contrastive learning approach using the sigmoid-softmax loss achieves the best performance across all metrics, with an F1-score of 0.69 and an accuracy of 69\%. This demonstrates the effectiveness of our learned embeddings in capturing the sector-level similarities and relationships between assets, despite only using returns data in the learning process. The individual sigmoid loss also performs well despite it's simplicity, with an F1-score of 0.68 and an accuracy of 68\%.
Interestingly, the aggregate sigmoid loss performs much worse than the other contrastive loss functions, suggesting that preserving the individual characteristics of positive and negative samples in the loss function is important for learning high-quality asset embeddings.

The proposed contrastive learning approach outperforms traditional time series baseline methods, as well as recent approaches for learning asset embedding from returns time series. This demonstrates the effectiveness of our framework in capturing meaningful relationships between financial assets.

\begin{figure}[t]
\centering
\includegraphics[width=0.45\textwidth]{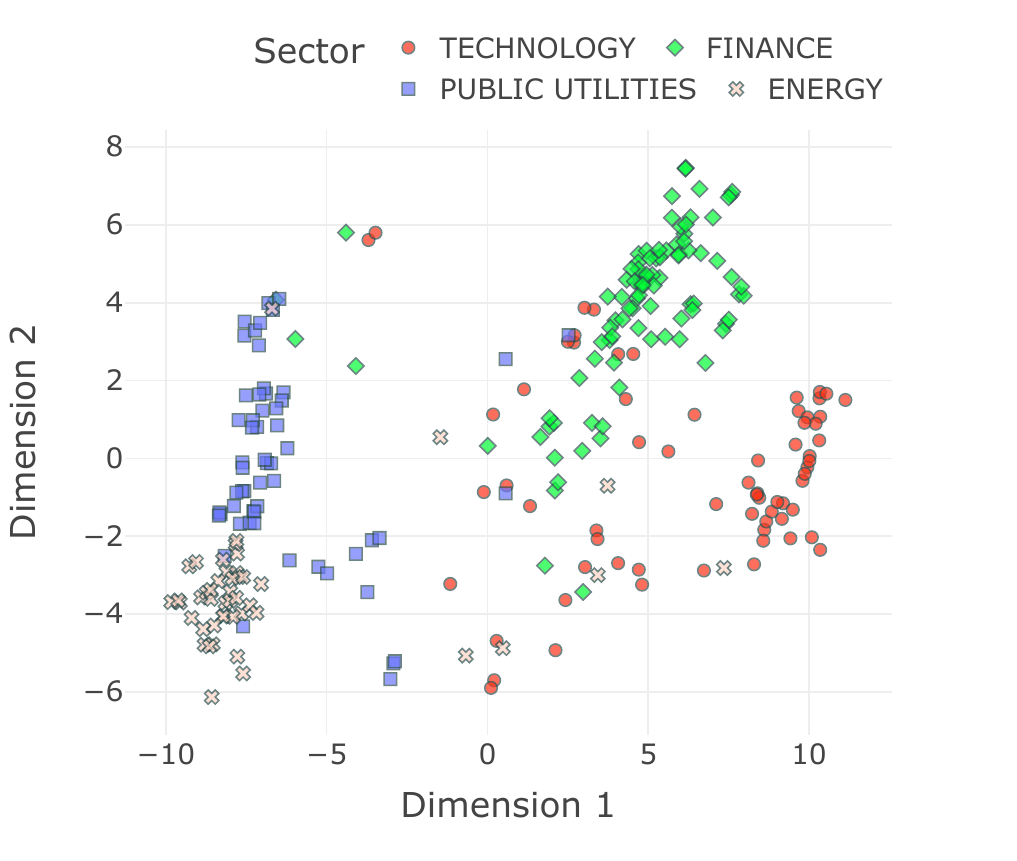}
\caption{2D visualization of the learned asset embeddings using t-SNE. Assets from the same industry sector tend to cluster together in the embedding space.}
\label{fig:contrastive:pca_2d}
\end{figure}

Figure \ref{fig:contrastive:pca_2d} visualizes the learned asset embeddings in two dimensions using t-SNE~\cite{van2008tSNE}, a dimensionality reduction technique that preserves the local structure of high-dimensional data. The visualization shows that assets from the same industry sector tend to cluster together in the embedding space, further validating the quality of our learned embeddings.

\subsubsection{Exploring the Learned Embedding Space}

\begin{table*}[]
\centering
\caption{Top 5 nearest neighbors for Pfizer Inc. (PFE) in the learned embedding space.\\Pfizer has ground truth sector and industry labels of \textit{Health Care} and \textit{Major Pharmaceuticals}.}
\label{tab:pfizer_neighbors}
\begin{tabular}{clllcc}
\toprule
\textbf{Rank} & \textbf{Company} & \textbf{Ticker} & \textbf{Sector} & \textbf{Industry} & \textbf{Similarity} \\
\midrule
1 & Eli Lilly and Company & LLY & Health Care & Major Pharmaceuticals & 0.9778 \\
2 & Baxter International Inc. & BAX & Health Care & Medical/Dental Instruments & 0.9718 \\
3 & Merck \& Company, Inc. & MRK & Health Care & Major Pharmaceuticals & 0.9706 \\
4 & Medtronic plc & MDT & Health Care & Biotechnology: Electromedica & 0.9620 \\
5 & Johnson \& Johnson & JNJ & Health Care & Major Pharmaceuticals & 0.9620 \\
\bottomrule
\end{tabular}
\end{table*}
To gain further insight into the information captured by the learned asset embeddings, we explore the nearest neighbors of specific assets in the embedding space. Table \ref{tab:pfizer_neighbors} presents the five nearest neighbors of Pfizer Inc. (PFE), a major pharmaceutical company, based on cosine similarity in the embedding space. We observe that the nearest neighbors are all classified as healthcare companies in the traditional sector labels, with the three of the top five belonging to the same finer grained industry classification, ``Major Pharmaceuticals'', as Pfizer. This example demonstrates that the embeddings have effectively captured the sector and industry-level relationships between assets.

Interestingly, the embeddings also capture nuanced relationships that may be missed by traditional sector classification schemes. Table \ref{tab:pg_mismatches} shows the five assets with the highest cosine similarity to Procter \& Gamble (PG), a company in the \textit{Basic Industries} sector and the ``Package Goods/Cosmetics'' industry. In this case, none of the nearest neighbours are from the same high-level Basic Industries sector, and they also span different industries such as ``Beverages'', ``Packaged Foods'', and ``Specialty Chemicals''. This suggests that the embeddings have learned similarities between assets that go beyond the rigid boundaries of traditional industry classifications.

\begin{table*}[]
\centering
\caption{High similarity ``mismatches'' for Procter \& Gamble (PG) in the learned embedding space.\\Procter \& Gamble has ground truth sector and industry labels of \textit{Basic Industries} and \textit{Package Goods/Cosmetics}.}
\label{tab:pg_mismatches}
\begin{tabular}{clllcc}
\toprule
\textbf{Rank} & \textbf{Company} & \textbf{Ticker} & \textbf{Sector} & \textbf{Industry} & \textbf{Similarity} \\
\midrule
1 & Colgate-Palmolive Company & CL & Consumer Non-Durables & Package Goods/Cosmetics & 0.9907 \\
2 & PepsiCo, Inc. & PEP & Consumer Non-Durables & Beverages (Production/Distri) & 0.9899 \\
3 & Conagra Brands, Inc. & CAG & Consumer Non-Durables & Packaged Foods & 0.9876 \\
4 & Clorox Company & CLX & Consumer Durables & Specialty Chemicals & 0.9832 \\
5 & Kimberly-Clark Corporation & KMB & Consumer Durables & Containers/Packaging & 0.9825\\
\bottomrule
\end{tabular}
\end{table*}

These examples demonstrate that our contrastive learning approach has captured meaningful relationships between assets that extend beyond traditional sector and industry classifications. By learning from the raw time series data in an entirely self-supervised manner, the embeddings are able to uncover nuanced similarities that may be overlooked by rigid classification schemes. This highlights the potential of our approach to provide new insights into the relationships between financial assets and to enable more data-driven decision making in various financial applications.

\subsection{Portfolio Optimization}
To further demonstrate the practical value of our learned asset embeddings, we consider a na\"ive hedging task designed to proxy portfolio optimization. We construct a simple long-only portfolio for each asset, where the asset is the target, and a single other asset is selected as the hedge based on the similarity information encoded in the embeddings (or the baseline similarity methods). For this experiment, we learn embeddings and compute similarity based on returns data from 2000-2012 and evaluate on the remaining period from 2013-2018.

For each target asset, we select the hedge asset randomly from the 25 most dissimilar assets (lowest cosine similarity in the embedding space) and repeat the experiment 100 times for robustness and to allow for statistical significance testing. With the selected hedge asset, we construct a two-asset portfolio with equal weights and evaluate its out-of-sample realized volatility over the test period. The hypothesis is that if the embedding space is learning information valuable for portfolio optimization applications, then we should see the embedding space yielding successful hedge assets that result in lower out-of-sample volatility.

We compare our approach with baseline similarity methods. For example, the most natural choice is to follow modern portfolio theory~\cite{markowitz1952portfolio} and select the hedge asset based on the lowest Pearson correlation with the target asset. We also compare against tailored similarity metrics~\cite{chun2020geometric} and recent embedding-based approaches~\cite{dolphin2022stock, dolphin2023iccbr, sarmah2022learning}. Table \ref{tab:contrastive:hedge_results} reports the average realized out-of-sample volatility across all portfolios for our contrastive learning approach with different loss functions and the baseline methods. The baseline methods are split between traditional similarity metrics and task-specific representation learning frameworks.

\begin{table}[t]
    \centering
    \caption{Portfolio hedging experiment results along with Tukey HSD test indicating significantly lower volatility than Pearson baseline at $\alpha=0.01$. * indicates the authors best effort implementation.}\label{tab:contrastive:hedge_results}
    \begin{tabular}{p{3.2cm}|cc}
        \toprule
        \textbf{Method} & \textbf{Avg Volatility} & \textbf{Significant} \\
        \midrule
        Pearson                             & 23.8\%                  & -                    \\
        Spearman                            & 24.0\%                  & \xmark                    \\
        \citet{chun2020geometric}\textsuperscript{*}                           & 23.9\%                  & \xmark                    \\
        \midrule
        \citet{dolphin2022stock}\textsuperscript{*} & 21.3\% & \cmark \\
        \citet{sarmah2022learning}\textsuperscript{*} & 22.0\% & \cmark \\
        \citet{dolphin2023iccbr}\textsuperscript{*} & 22.5\% & \cmark \\
        \midrule
        Individual Sigmoid                           & \textbf{19.1}\%                  & \cmark                    \\
        Aggregate Sigmoid                  & 20.3\%                  & \cmark                    \\
        Sigmoid-Softmax                     & 26.1\%                  & \xmark                    \\
        \bottomrule
    \end{tabular}
\end{table}

The results show that our contrastive learning approach with the individual sigmoid loss achieves the lowest average volatility of 19.1\%, outperforming the baseline method based on Pearson correlation with statistical significance. This indicates that the similarity information captured by our learned embeddings is effective in identifying good hedging pairs and constructing portfolios with lower risk.

The aggregate sigmoid loss also performs well, with an average volatility of 20.3\%, demonstrating its ability to capture overall characteristics of positive and negative samples for portfolio optimization. However, the hybrid sigmoid-softmax loss, despite its strong performance in sector classification, yields a higher average volatility compared to the baseline method, suggesting that the optimal choice of loss function may depend on the specific downstream task.

\begin{figure}[t]
\centering
\includegraphics[width=0.47\textwidth]{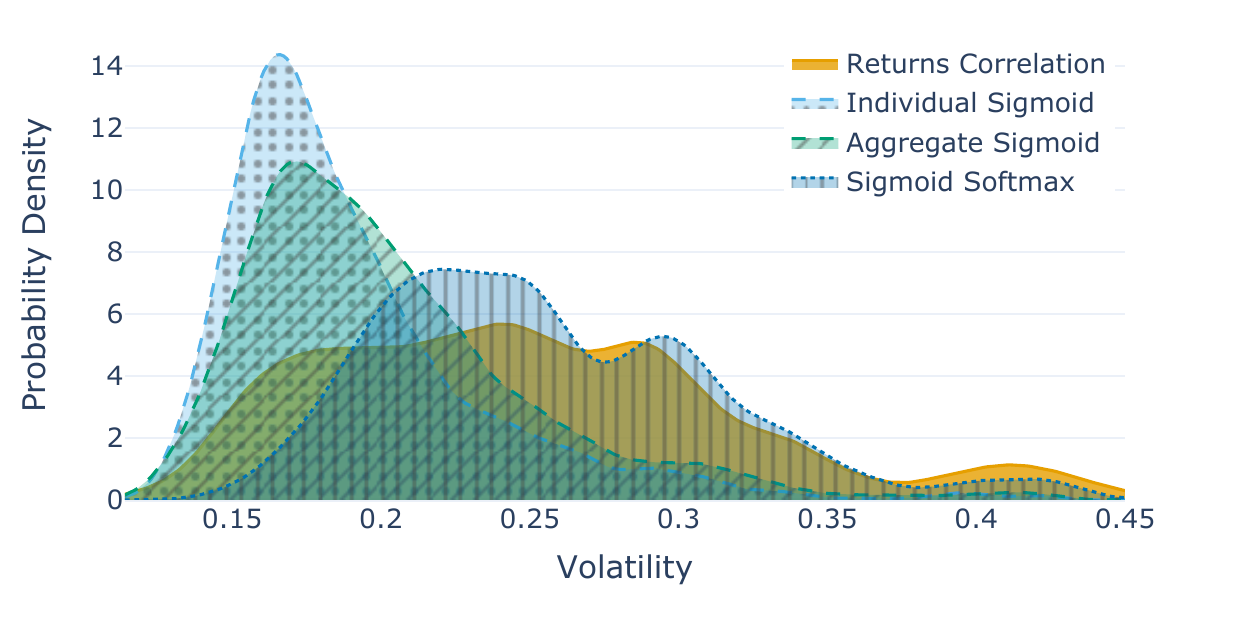}
\caption{Distribution of realized volatility for portfolios constructed using our individual sigmoid loss approach and the Pearson correlation baseline.}
\label{fig:contrastive:portfolio_opt_results}
\end{figure}

Figure \ref{fig:contrastive:portfolio_opt_results} presents the distribution of realized volatility for the portfolios constructed using the proposed embeddings and the Pearson correlation baseline. The portfolios based on our learned embeddings exhibit a lower median volatility and a tighter distribution compared to the baseline, further highlighting the effectiveness of our approach in capturing meaningful relationships between assets for portfolio optimization.

Overall, the experimental results on industry sector classification and portfolio optimization demonstrate the effectiveness of our contrastive learning framework in learning meaningful and informative asset embeddings from financial time series data. The superior performance compared to traditional methods and the practical value in downstream financial tasks highlight the potential of our approach in capturing complex relationships and similarities between financial assets.

\section{Discussion \& Conclusion}\label{sec:conclusion}
In this paper, we proposed a novel contrastive learning and sampling framework for learning embedding representations of financial assets from returns time series data. Our approach leverages the co-occurrence of assets in their returns similarity to generate informative positive and negative samples for the contrastive learning process. We introduced a statistical sampling strategy based on a hypothesis test of proportions to address the challenges associated with the noisy and stochastic nature of financial data. We explored various contrastive loss functions that capture the relationships between the anchor asset and the positive and negative samples in different ways, learning a discriminative and informative representation space.

Our experimental results on real-world financial datasets demonstrate the effectiveness of our learned asset embeddings in downstream tasks such as industry sector classification and portfolio optimization. The superior performance compared to popular time series classification techniques and tailored state-of-the-art approaches highlights the potential of our approach in capturing complex relationships and similarities between financial assets. The statistically significant benefits of our learned embeddings in the hedging experiment further emphasizes the potential of our framework to add value to real-world financial applications.

While our work focuses on financial time series data, the proposed techniques are also applicable to a wide range of non-financial time series. The ability to learn meaningful representations from raw time series data without relying on handcrafted features or domain-specific knowledge makes our approach highly adaptable to various application scenarios. Exploring the effectiveness of our contrastive learning framework in these non-financial domains is a promising avenue for future research.

One potential limitation of our approach is the scalability of the pairwise similarity computation over multiple windows, particularly for datasets with a large number of time series or very long time series. As the number of assets and the length of the time series grow, the computational complexity of calculating pairwise similarities across multiple windows can become intractable. However, this issue can be mitigated to some extent by adjusting the window size and stride parameters, effectively reducing the number of comparisons required. Furthermore, recent research has shown that this type of time series similarity calculation can scale without issue due to (i) the ability to leverage hardware and parallelize calculations and (ii) new algorithms that make the time complexity constant in subsequence length. For example, a recent paper demonstrates how to use GPUs to compute more than ten quadrillion pairwise comparisons for time series motifs and joins~\cite{zhu2018exploiting}. By implementing these advances in the similarity computation, we can overcome this limitation and ensure the scalability of our approach to larger datasets.

In conclusion, our work introduces a novel contrastive learning framework for learning asset embeddings from financial time series data, demonstrating its effectiveness in capturing meaningful relationships and similarities between assets. The proposed statistical sampling strategy and the exploration of various contrastive loss functions contribute to the robustness and discriminative power of the learned embeddings. The superior performance in industry sector classification and portfolio optimization highlights the practical value of our approach in real-world financial applications. We believe that our work opens up new avenues for representation learning in finance and has the potential to be extended to non-financial time series domains, paving the way for more accurate and data-driven decision-making in various fields.

%%
%% The acknowledgments section is defined using the "acks" environment
%% (and NOT an unnumbered section). This ensures the proper
%% identification of the section in the article metadata, and the
%% consistent spelling of the heading.
\begin{acks}
This publication has emanated from research conducted with the financial support of Science Foundation Ireland under Grant number 18/CRT/6183 and 12/RC/2289\_P2.
\end{acks}

%%
%% The next two lines define the bibliography style to be used, and
%% the bibliography file.
\bibliographystyle{ACM-Reference-Format}
\bibliography{sample-base}

\end{document}